\documentclass[conference]{IEEEtran}
\IEEEoverridecommandlockouts
\usepackage{cite}
\usepackage{amsmath,amssymb,amsfonts}
\usepackage{algorithmic}
\usepackage{graphicx}
\usepackage{textcomp}
\usepackage{xcolor}
\usepackage{lipsum}
\usepackage{cleveref}
\def\BibTeX{{\rm B\kern-.05em{\sc i\kern-.025em b}\kern-.08em
    T\kern-.1667em\lower.7ex\hbox{E}\kern-.125emX}}

\begin{document}
\title{On the Performance of Generative Adversarial Network (GAN) Variants: A Clinical Data Study}

\author{\IEEEauthorblockN{$^{\circ}$Jaesung Yoo, 
$^{\star}$Jeman Park, 
$^{\bullet}$An Wang, 
$^{\dag}$David Mohaisen, and $^{\circ,\ddag}$Joongheon Kim}
\IEEEauthorblockA{$^{\circ}$School of Electrical Engineering, Korea University, Seoul, Republic of Korea 
\\$^{\star}$School of Electrical and Computer Engineering, Georgia Institute of Technology, Atlanta, GA, USA
\\$^{\bullet}$Department of Electrical Engineering and Computer Science, Case Western Reserve University, Cleveland, OH, USA
\\$^{\dag}$Department of Computer Science, University of Central Florida, Orlando, FL, USA
\\$^{\ddag}$Artificial Intelligence Engineering Research Center, College of Engineering, Korea University, Seoul, Republic of Korea
\\
E-mails: \texttt{jsyoo61@korea.ac.kr}, 
\texttt{parkjeman122@gmail.com},
\texttt{axw474@case.edu}, \\ 
\texttt{mohaisen@ucf.edu},
\texttt{joongheon@korea.ac.kr}
}
}

\maketitle

\begin{abstract}
    Generative Adversarial Network (GAN) is a useful type of Neural Networks in various types of applications including generative models and feature extraction. Various types of GANs are being researched with different insights, resulting in a diverse family of GANs with a better performance in each generation. This review focuses on various GANs categorized by their common traits.
\end{abstract}


\section{Introduction}\label{sec:intro}

Deep learning has many applications in the area of medical informatics~\cite{JeonKKKMK19,icdcs18,emdl18,icufn17}. 
Generative adversarial network (GAN) is one of the most widely used deep learning models today which was first introduced in 2014 by \textit{Goodfellow et al}\cite{goodfellow2014generative}. The notable idea of GAN is the adversarial nets framework. This framework has two modules: generator and discriminator, which competes with adversarial objectives. It allows the generator to learn the distribution of the data through feedback from the discriminator. Since the advent of GAN, it has gained considerable popularity and numerous variants of GAN are being proposed continuously.
The variants of GAN are largely divided into two main types. The first type copes with the instability in the learning process of vanilla GAN. These variants approached the learning problems with various insights, resulting in diverse loss functions to achieve better performance. The second type is specialized in each task (e.g., image-to-image translation \cite{choi2018stargan} or super-resolution \cite{ledig2017photo}). These variants construct their model architectures differently than vanilla GAN.

In this paper, we first analyze the mechanism and chronic learning problems of the vanilla GAN. Then, we introduce some variants of GAN in the perspectives of loss functions and model architectures. We then conduct experiments on surgical data augmentation using different GAN types to compare the performance.

\section{Analysis of the Vanilla GAN}\label{sec:GAN}

The vanilla GAN simply consists of a generator model $G$ and its adversarial discriminator model $D$. First, $D$ receives fake data which are generated from $G$'s model distribution, or real data from the real data distribution. Then $D$ is trained to precisely distinguish these real and fake data. Next, $G$ generates and sends fake data to $D$. $D$ evaluates how close the fake data is to the real and this result is fed back to $G$. Finally, $G$ is trained as to confuse $D$ from differentiating data. This process can be described with the value function of the following:

\begin{multline}
    \label{eq:GAN_value}
    \min_{\textit{G}}\max_{\textit{D}} V(D, G) = \\
    \mathop{\mathbb{E}}[\log(D(x))] + \mathop{\mathbb{E}}[\log(1- D(G(z)))]
\end{multline}
where $x$ is a sample from real data distribution and $z$ is the input noise for the generator $G$. Thus the loss function can be described as the following:
\begin{align}
    \label{eq:loss_D}
    \mathcal{L}_{\textrm{D}} &=
    - \mathop{\mathbb{E}}[\log(D(x))] - 
    \mathop{\mathbb{E}}[\log(1-D(G(z)))] \\
    \label{eq:loss_G1}
    \mathcal{L}_{\textrm{G}} &= \mathop{\mathbb{E}}[\log(1-D(G(z)))].
\end{align}

For the sake of better convergence, the following loss function could also be used as it proposes a steeper gradient towards undesired values.
\begin{align}
    \label{eq:loss_G2}
    \mathcal{L}_{\textrm{G}} &= -\mathop{\mathbb{E}}[\log(D(G(z)))]
\end{align}

This two-player minimax game is proceeded with two adversarial gradient descent algorithms which optimizes the loss function in the opposite direction. This may look straightforward but it has some problems with learning instability as follows:

\subsubsection{Vanishing gradient}
If the discriminator outperforms, the generator can fail to learn because the gradient vanishes\cite{arjovsky2017towards}. In other words, optimal discriminator does not provide sufficient feedback for the generator to learn properly.

\subsubsection{Mode collapse}
The desired generator should produce a variety of outputs that resembles the overall data distribution. However, it sometimes generates only a fraction (mode) of the overall data distribution, resulting in a specific mode that still minimizes the adversarial loss function. In response to this phenomenon, the discriminator might figure out the specific mode of the generator and adjust its weight to criticize the generator. This triggers the generator to move its output distribution to the next fraction of the real data distribution, leading to oscillating modes of data distribution that might not converge\cite{metz2016unrolled}.

\subsubsection{Non-convergence}
As GAN is based on a minimax game based on an adversarial loss function, the generator and the discriminator are under a never-ending loop of oscillation. When the generator improves enough to fool the discriminator, the discriminator would have around 50\% accuracy. At this point, the discriminator cannot give appropriate feedback and rather gives random feedback which degrades the generator's performance again. After the generator's performance gets worse, the generator will learn the appropriate features again and this oscillation continues when both discriminator and generator jointly search for equilibrium.

Many GAN variants were proposed to remedy these learning problems through different adjustments such as loss modifications or novel model architectures.

\section{Loss based GAN}\label{sec:lossbasedGAN}
One possible way of stabilizing the training of GAN is to modify its loss functions. 

Since loss functions provide guidance for model weights to follow in the vast state space, it is important that the functions faithfully represent the ultimate goals of the optimization problem.

\subsection{Least Squares GAN (LSGAN)}
LSGAN argued that the sigmoid cross entropy loss function would lead to the problem of vanishing gradients when updating the generator using the fake samples which are on the correct side of the GAN's decision boundary, but are still far from the real data\cite{mao2017least}. Because these samples are in the high confidence area of the decision boundary, so there is almost no loss in the vanilla GAN. However, these samples are isolated from real data, so they seem unrealistic. To remedy this issue, LSGAN introduced the least-squares loss function for pulling them close to the decision boundary. The loss functions are as the following:
\begin{eqnarray}
    \label{eq:LSGAN_loss}
    \min_{\textit{D}} V(D) &=& \frac{1}{2}\mathop{\mathbb{E}}[(D(x) - b)^{2}] + \nonumber \\ 
    & & \frac{1}{2}\mathop{\mathbb{E}}[(D(G(z)) - a)^{2}] \\
    \min_{\textit{G}} V(G) &=& \frac{1}{2}\mathop{\mathbb{E}}[(D(G(z)) - c)^{2}]
\end{eqnarray}
where $a$ and $b$ are the labels for fake data and real data, and $c$ denotes the value that $G$ wants $D$ to believe for fake data, respectively. By these loss functions, LSGAN will penalize the samples which are far from the real data even though they are correctly classified, and pull them to the decision boundary. If a GAN has been successfully learned, its decision boundary is formed to pass through the manifold of the real data. Accordingly, pulling the samples to the decision boundary makes them be closer to the manifold, allowing $G$ to generate more realistic data.

\subsection{Wasserstein GAN (WGAN)}
WGAN uses Wasserstein distance to stabilize training\cite{arjovsky2017wasserstein}. Along with Kullback-Leibler divergence (KL-divergence) and Jensen–Shannon divergence(JS-divergence), Wasserstein distance is a metric to measure the distance between two probability distributions. A common drawback of KL-divergence and JS-divergence is that when two distributions are far different, the loss function becomes flat, resulting in a vanishing gradient problem. In applications, such as GAN, where the two modules compete fiercely, it is easy for the two modules to become far apart in the mean probability distribution, where the vanishing gradient happens and leads to non-convergence. To address this problem, Wasserstein distance can be used as its gradient does not vanish. Wasserstein distance is defined as the following:
\begin{equation}
    \label{eq:Wdistance}
    \begin{aligned}
        \mathcal{L}_{\textrm{WGAN}} = ||D(x) - D(G(z))||.
    \end{aligned}
\end{equation}
Unlike KL-divergence, the Wasserstein distance does not have logs which make it fairly linear. This can lead to more stable training than using KL-divergence.

\section{Architecture based GAN}\label{sec:archbasedGAN}
GAN can also be improved with additional variants on architecture. The following Neural Networks are variants of GAN resulting from the following insights: Generator capable of having Variational Autoencoder (VAE) architecture and Discriminator having multiple outputs.

\subsection{Variational Autoencoder GAN (VAEGAN)}
VAEGAN has its simple generator substituted with Variational Autoencoder (VAE)\cite{larsen2016autoencoding}. Thus, it inherits the benefits of GAN and VAE. The VAE acts as a generator and creates augmented data while the discriminator tries to score them. This structure enhances the power of feature extraction. In the training process, the VAE structure makes learning stable. After training completes, VAEGAN is more effective than GAN to generate new samples since passing a random sampled latent vector results in new augmented samples.

VAE has two submodules: Encoder $q$ and a Decoder $G'$. The encoder encodes the given data into a latent variable $z$. The decoder then reconstructs the data from the latent variable $z$. Thus reconstruction loss is applied as:
\begin{equation}
    \label{eq:loss_rec}
        \mathcal{L}_{\textrm{rec}} = -\mathop{\mathbb{E}}[\log G'_{\hat{z}\sim q(x)}(x|\hat{z})]
\end{equation}
where $\hat{z}$ is sampled from the distribution $q(x)$. The loss functions for VAEGAN is calculated as the linear combination of GAN loss and VAE reconstruction loss, with one possible additional loss. Kullback-Leibler divergence (KL-divergence) is optionally applied to the latent variable to reduce model complexity. The latent variable $z$ is conventionally restrained to $N(0,1)$. The KL-divergence of two Gaussians is as the following:
\begin{eqnarray}
    \label{eq:KLDivergence}
        \mathcal{L}_{\textrm{KLD}} &=& D_{\textrm{KL}}(q(z|x)||p(z)) \nonumber \\
        &=& \log\left(\frac{\sigma_p}{\sigma_q}\right)+
        \frac{\sigma_q^2+(\mu_q-\mu_p)^2}{2\mu_p^2} - \frac{1}{2}
\end{eqnarray}
where $p$ denotes the prior distribution of the latent space, which is conventionally assumed to be $N(0,1)$. Thus, $\mu_2=0$, $\sigma_2=1$. Since VAE acts as a generator, we can define $G$ in a way to utilize the same loss function from \eqref{eq:loss_G2}
\begin{equation}
    \label{eq:G_VAE}
        G(z) = G'_{\hat{z}\sim q(x)}(\hat{z}).
\end{equation}

Integrating the losses from GAN and VAE, the final loss for the VAE becomes:
\begin{eqnarray}
    \label{eq:loss_VAE}
        \mathcal{L}_{\textrm{VAE}} &=& 
        \mathcal{L}_{\textrm{G}} + \lambda_{\textrm{rec}} \mathcal{L}_{\textrm{rec}} + \lambda_{\textrm{KLD}}\mathcal{L}_{\textrm{KLD}}
\end{eqnarray}
where $\lambda_{\textrm{rec}}$, $\lambda_{\textrm{KLD}}$ are hyperparameters which denote the weights of the losses. The loss for the discriminator is the same as Equation~\eqref{eq:loss_D}

\subsection{Auxiliary Classifier GAN (ACGAN)}
ACGAN is a variant of GAN in which the discriminator has a classifier output along with the standard True/False output\cite{odena2017conditional}. Since the discriminator can also act as a classifier, the generator can be modified to accept a condition vector to produce samples of a different class for the sake of versatility and functionality. Thus, ACGAN is appropriate in applications which has categorical distributions. In the case of ACGAN, the generator is noted as $G(c,z)$ where $c$ denotes the class vector which acts as a condition vector. The discriminator is noted as $D_\textrm{S}(x)$, $D_\textrm{C}(c,x)$ which are outputs from the discriminator that distinguishes True/False data and classifies categories respectively.

Loss functions of ACGAN can be divided into 2 parts: distinguishing the source and the class. The loss function determined from the source of the data is the same as~\Cref{eq:loss_D,eq:loss_G2}:
\begin{align}
    \label{eq:loss_D_S}
    \mathcal{L}_{\textrm{D}_\textrm{S}} &= -\mathop{\mathbb{E}}[\log(D_\textrm{S}(x))] - 
    \mathop{\mathbb{E}}[\log(1-D_\textrm{S}(G(c,z)))] \\
    \label{eq:loss_G_S}
    \mathcal{L}_{\textrm{G}_\textrm{S}} &= -\mathop{\mathbb{E}}[\log(D_\textrm{S}(G(c,z)))]
\end{align}
The loss function from the class of the data is defined in a way to encourage cooperation between the generator and the discriminator, rather than operating in adversary.
\begin{align}
    \label{eq:loss_D_C}
    \mathcal{L}_{\textrm{D}_\textrm{C}} &= 
    - \mathop{\mathbb{E}}[\log(D_\textrm{C}(c,x))] - 
    \mathop{\mathbb{E}}[\log(D_\textrm{C}(c,G(c,z)))] \\
    \label{eq:loss_G_C}
    \mathcal{L}_{\textrm{G}_\textrm{C}} &= - \mathop{\mathbb{E}}[\log(D_\textrm{C}(c,G(c,z)))]
\end{align}
The final loss functions for the generator and the discriminator is the sum of the two loss functions defined in \Crefrange{eq:loss_D_S}{eq:loss_G_C}.
\begin{align}
    \label{eq:ACGAN_loss_D}
    \mathcal{L}_{\textrm{D}} &= \mathcal{L}_{\textrm{D}_\textrm{S}} + \lambda_{\textrm{D}_\textrm{C}} \mathcal{L}_{\textrm{D}_\textrm{C}} \\
    \label{eq:ACGAN_loss_G}
    \mathcal{L}_{\textrm{G}} &= \mathcal{L}_{\textrm{G}_\textrm{S}} +
    \lambda_{\textrm{G}_\textrm{C}} \mathcal{L}_{\textrm{G}_\textrm{C}}
\end{align}
where $\lambda_{\textrm{D}_\textrm{C}}$, $\lambda_{\textrm{G}_\textrm{C}}$ are hyperparameters which denote the weights of the losses.

In the case where the class label is not available, variants of ACGAN can be introduced by setting the class loss function with entropy rather than categorical cross-entropy. This ensures that the model learns to predict in the absence of a class label. In applications which the generator wants to fool the discriminator, the entropy class loss function can be maximized rather than minimized.

\subsection{Auxiliary Classifier Variational Autoencoder (ACVAE)}
Motivated by VAEGAN and ACGAN, both generator and discriminator architecture can be modified for improvement. ACVAE uses Auxiliary Classifier (AC) as the discriminator and VAE as the generator~\cite{kameoka2019acvae}. The AC structure allows the generator to create samples of various classes while the VAE structure allows stable training and flexible generation of samples. The generator of ACVAE is a VAE structure which receives a class vector $c$ as condition. Since there are two submodules in VAE: Encoder $q$ and a Decoder $G'$, there are 3 possible types of generator structures that receives the condition vector. The case which only the Encoder $q$ receives $c$, case which only the Decoder $G'$ receives $c$, and the case which both submodules receives $c$. The appropriate structure depends on the applications. When the last type of architecture is assumed, the encoder is noted as $q(c,x)$ and the decoder $G'(c,z)$.

Since the previous two insights on architecture are used in ACVAE the corresponding loss functions are also followed. The generator $G$ can be redefined to utilize the same loss function from \eqref{eq:ACGAN_loss_D},\eqref{eq:ACGAN_loss_G}
\begin{eqnarray}
    \label{eq:ACVAE_G}
    G(c,z) = G'_{\hat{z} \sim q(c,x)}(c,\hat{z}).
\end{eqnarray}

The loss for the VAE is identical to \eqref{eq:loss_VAE}, and the loss for the discriminator is identical to \eqref{eq:ACGAN_loss_D}
\begin{eqnarray}
    \label{eq:ACVAE_loss_VAE}
    \mathcal{L}_{\textrm{VAE}} &=& 
        \mathcal{L}_{\textrm{G}} + \lambda_{\textrm{rec}} \mathcal{L}_{\textrm{rec}} + \lambda_{\textrm{KLD}}\mathcal{L}_{\textrm{KLD}} \nonumber \\ 
        &=& \mathcal{L}_{\textrm{G}_\textrm{S}} +
    \lambda_{\textrm{G}_\textrm{C}} \mathcal{L}_{\textrm{G}_\textrm{C}} + \lambda_{\textrm{rec}} \mathcal{L}_{\textrm{rec}} + \lambda_{\textrm{KLD}}\mathcal{L}_{\textrm{KLD}}.
\end{eqnarray}


\begin{figure*}[ht!]
\centering
\setlength{\tabcolsep}{2pt}
\renewcommand{\arraystretch}{0.2}
\begin{tabular}{cc}
\includegraphics[width=0.4\textwidth]{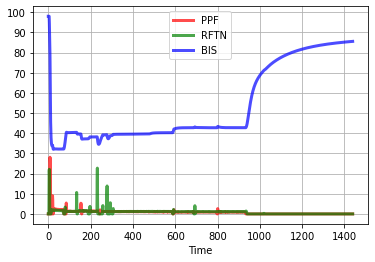} &
\includegraphics[width=0.4\textwidth]{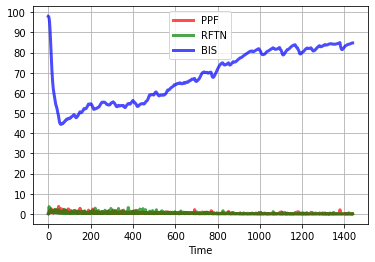}
\tabularnewline
(a) Ground truth & (b) Vanilla GAN
\end{tabular}

\vspace{0.2in}

\begin{tabular}{ccc}
\includegraphics[width=0.33\textwidth]{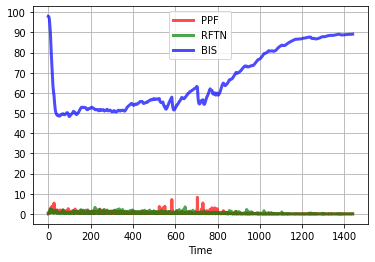} &
\includegraphics[width=0.33\textwidth]{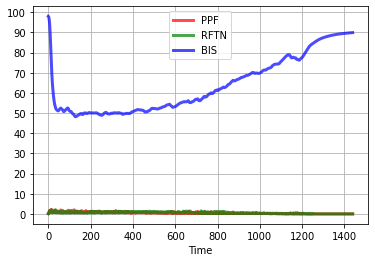} &
\includegraphics[width=0.33\textwidth]{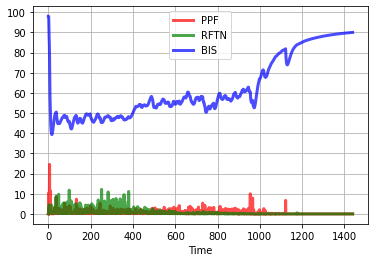}
\tabularnewline
(c) LSGAN & (d) WGAN & (e) VAEGAN
\tabularnewline
\end{tabular}

\vspace{0.1in}

\caption{Examples of Data Augmentation result using different GAN variants. For PPF and RFTN the $y$-axis equals dose/10sec, and for BIS the $y$-axis is dimensionless. ACGAN and ACVAE are not included as the application does not have classes.}
\label{fig:results}
\end{figure*}

\section{Experiments}\label{sec:exp}

While machine learning requires a vast amount of data, surgical databases are limited in quantity due to practical constraints. Thus, data augmentation techniques are used in medical applications to aid the learning process of the machine learning model\cite{frid2018gan, shin2018medical}.

Data augmentation experiment using anesthetic surgical data was performed to compare the results of different GAN variants. The data includes four components which consist of 2 types of anesthetic drug dosage history: Propofol(PPF) and Remifentanil(RFTN), the anesthetized response of the patient: Bispectral Index(BIS), and the covariates of the patient. Lower BIS indicates a deeply anesthetized state and BIS around 50 is desired. Data augmentation was performed by the generator which receives patient covariates and Gaussian noise as input and creates fabricated drug dosage history. The augmented drug dosage was put into the pharmacokinetic-pharmacodynamic model (PK-PD model), a traditional patient response model to compute the BIS response of the augmented data\cite{short2016refining}. The BIS response to augmented drug dosage was used to monitor the training process of GAN.

Fig.~\ref{fig:results} is the examples of augmented data using different types of GAN. Real surgical data tends to have a high peak at the start of the surgery, which occurs to fully anesthetize the patient. At the end of the surgery, the drug dosage is low to awake the patient. From Fig.~\ref{fig:results} it can be seen that vanilla GAN succeeds in generating synthetic surgical data which ensures a roughly similar BIS response. However, the drug dosages are distributed all along with the timestamps, unlike the ground truth surgical data which has certain peaks at certain points. Due to distributed drug dosage, the BIS response to the augmented drug dosage of vanilla GAN keeps increasing. Results of LSGAN and WGAN are roughly similar and the performance is better than vanilla GAN as the BIS response to the augmented data better stays around 50. VAEGAN shows the best result that resembles the ground truth data, with sharp peaks upfront and BIS around 50. However, this does not ensure that VAEGAN always outperforms the others since deep learning architectures are application-dependent. Variants of GAN are improved versions of vanilla GAN, but the performance of each model is application-dependent.

\section{Concluding Remarks}\label{sec:con}
GAN is a powerful structure that invigorates feature extraction but is unstable along with its uncertainty of convergence during training. Its adversarial insight is the core motivation of model improvement but its vanilla structure is too simple and has numerous unknown factors affecting its results which accounts for GAN's instability. Thus various attempts to further restrict the search space of GAN prevails and it is an ongoing matter that needs more research.

The human brain is evidently a massive modular neural network. Since nature and evolution have divided the brains of animals into numerous sub-modules rather than to operate as a whole, it can be boldly deduced that operating with the harmony of sub-modules performs better than a single large module. GAN is a powerful type of modular neural network, and the modularizing neural network would be one of the main keys to stabilizing large deep neural networks that perform powerful tasks.

\section*{Acknowledgement}\label{sec:ack}
This research was supported by the MHW (Ministry of Health and Welfare), Korea (Grant number: HI19C0842) supervised by the KHIDI (Korea Health Industry Development Institute) and the project title is DisTIL (Development of Privacy-Reinforcing Distributed Transfer-Iterative Learning Algorithm). All authors in this paper have equal contributions (first authors). J. Kim is the corresponding author of this paper.



\end{document}